\documentclass{article}
\usepackage{spconf,amsmath,graphicx}

\usepackage{multirow}
\usepackage{booktabs}
\usepackage{tabularx}
\usepackage{ragged2e}
\usepackage{siunitx}
\usepackage{float}
\usepackage{tikz}
\usepackage{pgfplots}
\usepackage{placeins}

\title{Efficiently Enhancing General Agents with Hierarchical-Categorical Memory}
\name{Changze Qiao\textsuperscript{1}, Mingming Lu\textsuperscript{1,*}\thanks{*Corresponding author}}
\address{\textsuperscript{1}School of Computer Science and Engineering, Central South University}

\begin{document}
%
\maketitle
\begin{abstract}

With large language models (LLMs) demonstrating remarkable capabilities, there has been a surge in research on leveraging LLMs to build general-purpose multi-modal agents. However, existing approaches either rely on computationally expensive end-to-end training using large-scale multi-modal data or adopt tool-use methods that lack the ability to continuously learn and adapt to new environments. In this paper, we introduce EHC, a general agent capable of learning without parameter updates. EHC consists of a Hierarchical Memory Retrieval (HMR) module and a Task-Category Oriented Experience Learning (TOEL) module. The HMR module facilitates rapid retrieval of relevant memories and continuously stores new information without being constrained by memory capacity. The TOEL module enhances the agent’s comprehension of various task characteristics by classifying experiences and extracting patterns across different categories. Extensive experiments conducted on multiple standard datasets demonstrate that EHC outperforms existing methods, achieving state-of-the-art performance and underscoring its effectiveness as a general agent for handling complex multi-modal tasks.

\end{abstract}
\section{Introduction}
\label{sec:intro}
The field of artificial intelligence has long been dedicated to creating general-purpose intelligent assistants \cite{li2024multimodal} capable of following multi-modal user instructions and efficiently accomplishing various real-world tasks. With large language models (LLMs) demonstrating remarkable capabilities, there has been a proliferation of research focused on leveraging LLMs to construct general-purpose multi-modal assistants.

Current approaches to building multi-modal agents primarily fall into two categories \cite{li2024multimodal}. On the one hand, end-to-end training integrates LLMs directly with multi-modal data. Researchers collect large-scale image-text datasets and multi-modal instruction-following data to continually train LLMs, equipping them with visual information-processing abilities. Several models \cite{zhao2023mmicl, li2023otter, nguyen2022coarse, bai2023qwen} have demonstrated impressive capabilities in visual understanding and reasoning. However, these methods are computationally expensive, require direct access to LLM parameters, and may restrict the flexibility and generalization capabilities of the LLMs. On the other hand, tool-use approaches enable LLMs to invoke different tools to accomplish required (sub)tasks through carefully designed prompts without additional model training. Notable examples include  VisProg \cite{gupta2023visual}, ViperGPT \cite{suris2023vipergpt}, and Visual ChatGPT \cite{wu2023visual}. Such methods efficiently perform a wide range of visual tasks using corresponding tools, offering a cost-effective integration into AI agents.

However, most existing tool-use methods do not enable agents to acquire new knowledge, thus limiting their continuous learning and adaptation to new environments. AssistGPT \cite{gao2023assistgpt} updates LLMs post-deployment through in-context learning, and CLOVA \cite{gao2024clova} updates both LLMs and visual tools during reflection and learning phases. Prompt-based methods can enhance the sequential decision-making and planning capabilities of LLMs by providing a few in-context examples \cite{hao2023reasoning}. Nevertheless, due to the context-window limitations of LLMs \cite{packer2023memgpt}, these agents cannot recall previously encountered information, restricting their learning capability beyond a limited number of examples. ExpeL \cite{zhao2024expel} addresses this by proposing experiential learning, where agents autonomously collect experiences from training tasks through trial-and-error, distill natural-language insights from these experiences, and reuse successful experiences as in-context examples at test time. Yet, as memory size grows and the need for semantic structuring increases, these systems encounter memory redundancy and overhead issues. Additionally, diverse task types stored in memory can interfere with the outputs of LLMs, making it essential for agents to focus specifically on relevant task types when handling various tasks. HAMMR \cite{castrejon2024hammr} targets multiple types of visual question-answering (VQA) problems by constructing a hierarchical multi-modal system that allows a high-level agent to invoke low-level agents specialized in specific task types. Despite achieving promising results, HAMMR does not address a broader variety of task types in more general scenarios, and its multi-layer agent structure is relatively complex.

In this research, we present EHC, a novel general-purpose agent framework designed to address the aforementioned challenges. To mitigate memory redundancy and overhead, we introduce the Hierarchical Memory Retrieval (HMR) module. Inspired by traditional operating system caching mechanisms, HMR adopts a dual-pool architecture consisting of a Fast-Access Memory Pool (in RAM) and a Deep-Retrieval Memory Pool (in an external database). During agent-environment interactions, HMR employs a dynamic migration strategy to manage memory flow between the two pools. This design enables rapid retrieval of relevant memories while significantly reducing storage overhead, effectively supporting continuous learning in open environments. To handle diverse multi-modal tasks, we propose the Task-Oriented Experiential Learning (TOEL) module, which aims to balance flexibility and explainability in the learning process. TOEL continuously collects experiences from agent-environment interactions without requiring additional annotation. By combining predefined categories with LLM-based few-shot reasoning, TOEL classifies experiences, extracts category-specific patterns, and distills actionable knowledge to guide task execution. This allows agents to better understand the characteristics of various task types, formulate more effective solutions, and significantly enhance their adaptability and problem-solving capabilities across tasks. Moreover, clear experience classification improves system transparency and explainability, making the agent’s decision-making process more understandable and easier to debug—crucial for the improvement and optimization of complex systems.

\begin{figure*}[htb]

\centering
\centerline{\includegraphics[width=17cm, height = 7cm]{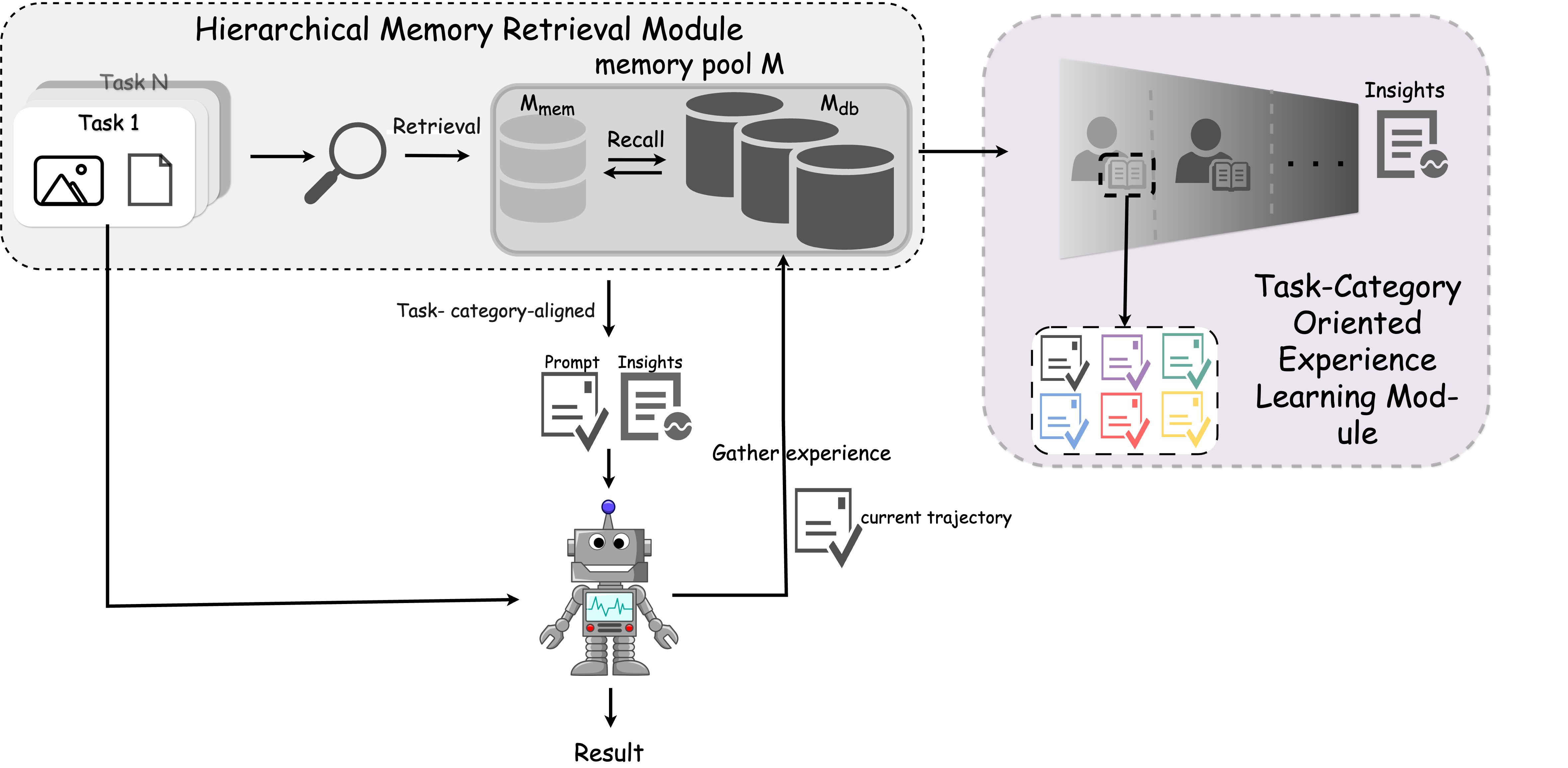}}
\caption{The framework of EHC.}
\label{EHC}
\end{figure*}

\section{METHODOLOGY}
\label{sec:pagestyle}

The proposed EHC architecture, as illustrated in Fig.~\ref{EHC}, comprises two core components: the Hierarchical Memory Retrieval (HMR) module and the Task-Type Oriented Experience Learning (TOEL) module. These modules work collaboratively to establish a well-structured and semantically enriched memory system, which not only facilitates efficient information storage and retrieval but also enables the agent to learn patterns across different memory categories.

\subsection{Hierarchical Memory Retrieval Module}

Formally, given a memory set $M$, frequently accessed or recently used memories $M_s$ are maintained in the fast-access pool $M_{\text{mem}}$, which has a fixed capacity $C$, while other important memories $M_r$ are stored in the scalable external database $M_{\text{db}}$.

During agent-environment interactions, memory transfers between $M_{\text{mem}}$ and $M_{\text{db}}$ are governed by HMR’s dynamic migration policy. When storing new observations, the system first attempts to cache them in $M_{\text{mem}}$. If the memory exceeds the capacity threshold $C$, the system migrates the least recently used $C/2$ entries to $M_{\text{db}}$ following a Least Recently Used (LRU) eviction policy.
For memory retrieval, the system first queries the fast-access pool using a similarity metric to retrieve the top-$k$ most relevant in-context examples (i.e., those with confidence scores greater than $\theta$). If the number of suitable matches is insufficient, the remaining candidates are retrieved from the deep-retrieval pool via database queries. Finally, the most relevant trajectories are integrated into the current context to generate the final output.

\subsection{Task-Category Oriented Experience Learning Module}

\textbf{Experience Collection.}  
We build upon prior work by collecting experiences throughout the agent’s operation, thereby eliminating the need for additional annotations. Each task is allotted a maximum of $T$ attempts. For the $i$-th task with content $c_i$, historical memory $h_i$ (retrieved from the hierarchical memory pool), current trajectory $p_0$, and initial reflections $r_0 = \emptyset$, the agent is prompted to generate a trajectory $p_t$ through combinatory reasoning for the $t$-th attempt. If the attempt succeeds, $p_t$ is stored in memory as a success experience. Otherwise, the agent performs self-reflection to obtain $r_{t+1} = \text{concat}(r_t, \text{LLM}(\cdot \mid p_t, c_i, h_i))$. If the maximum number of attempts is reached, $p_t$ and $r_{t+1}$ are stored as a failure experience; otherwise, the process repeats.

\noindent\textbf{Experience Classification.}  
To maintain both flexibility and interpretability, we combine predefined task categories with LLM-based few-shot inference. Based on domain analysis and task objective decomposition, we define a set of mutually exclusive and collectively exhaustive categories $C = \{c_1, c_2, \ldots, c_K\}$. For each collected experience, the LLM generates a candidate label $\hat{c}_i$. To mitigate LLM prediction instability, we compute semantic embeddings $\phi(\hat{c}_i)$ using BERT and compare them with embeddings $\phi(c_k)$ for each predefined category. The final category is assigned based on the highest cosine similarity:

\[
c_i = \arg\max_{c_k \in C} \text{sim}(\phi(\hat{c}_i), \phi(c_k)),
\]

where $\text{sim}(a, b)$ denotes cosine similarity. The memory pool is then organized into $M = \{M_{c_1}, M_{c_2}, \ldots, M_{c_K}\}$, where each $M_{c_k}$ stores success and failure trajectories corresponding to category $c_k$.

\noindent\textbf{Experience Learning.}  
Our learning strategy focuses on identifying patterns across experience categories. For each category $c_k$, the memory $M_{c_k}$ is split into intra-category success-failure pairs and cross-category comparison groups. Intra-category contrast pairs $C^{\text{fail/success}}_{\text{compare}}$ are formed by extracting fixed-length segments from success trajectories $M_{c_k^+}$, aiding in the identification of key success factors and common failure modes. Cross-category comparison groups $C^{\text{category}}_{\text{compare}}$ are formed by pairing success trajectories of $c_k$ with those from other categories $c_{\text{non-}k}$, revealing inherent differences among task types. During insight generation, the insight set is initialized as $\hat{\iota} = \emptyset$, and the LLM is iteratively prompted with memory data to generate new insights. These insights $\hat{\iota}_{c_k}$ can be updated through operations such as \texttt{ADD}, \texttt{EDIT}, \texttt{UPVOTE}, and \texttt{DOWNVOTE}. Each newly added insight starts with an initial weight, which increases or decreases based on subsequent feedback. When an insight’s weight drops to zero, it is removed from the pool. Finally, the EHC framework leverages the generated insight set $\hat{\iota}$ during the inference phase.

\subsection{Inference Based on Task Category}

During the inference phase, EHC integrates an information-rich memory pool $M$, consisting of trajectories from completed tasks and insights obtained through experiential learning. For program generation, EHC adopts the compositional visual reasoning framework \cite{gupta2023visual}. What distinguishes our approach is that, for each specific task, EHC retrieves insights of the same category from the memory pool and selects the top-$k$ most relevant trajectories within that category as few-shot contextual examples. These, combined with a trajectory template, are provided to the LLM to guide generation. For qualitative results, refer to Section~3.4.

\section{EXPERIMENT}

\subsection{Experimental Setup}

\textbf{Datasets and Evaluation Protocol.}  
To evaluate EHC, we conducted experiments using standard benchmark datasets and widely adopted evaluation metrics. In alignment with recent state-of-the-art studies, we used the GQA \cite{hudson2019gqa} and NLVR2 \cite{suhr2018corpus} datasets for compositional visual question answering (VQA) and multi-image reasoning, reporting top-1 accuracy. For factual grounding and referring expression comprehension, we used RefCOCO, RefCOCO+, and RefCOCOg \cite{yu2016modeling, kazemzadeh2014referitgame}, evaluated using Intersection over Union (IoU). For language-guided image editing, we adopted the MagicBrush \cite{zhang2023magicbrush} dataset, with performance measured by the CLIP-I score.

\noindent\textbf{Implementation Details.}  
We employed CLOVA \cite{gao2024clova} as the baseline and utilized its publicly available toolkit. We predefined several instruction categories---\textit{judgment}, \textit{counting}, \textit{recognition}, \textit{comparison}, \textit{addition}, \textit{removal}, and \textit{replacement}---and initialized both internal and external memory pools with five contextual examples per category.

\subsection{Main Results}

To assess the performance of our model, we compared EHC with several competitive baselines, including LLM-based agents and end-to-end models. The results, summarized in Table.~\ref{main result}, show that EHC significantly outperforms the baseline, achieving a 7.8\% and 4.0\% improvement in accuracy on the GQA and NLVR2 datasets, respectively. These gains underscore the effectiveness of the HMR and TOEL modules in managing complex multimodal tasks. The relatively smaller improvement observed on the image tagging task may be attributed to its simplicity and limited instruction diversity.

EHC also consistently outperforms other tool-use methods, further demonstrating its robustness. Compared to end-to-end models, EHC substantially narrows the performance gap. While end-to-end models typically rely on large-scale multimodal data for training and fine-tuning---constraining their flexibility and generalization---EHC’s structured memory management and task-oriented learning approach allow it to process diverse multimodal tasks more efficiently, without incurring the heavy computational costs and resource demands associated with end-to-end training.

\begin{table}[htb]
    \caption{\textbf{Comparison of results with previous models.}} 
    \label{main result}
    \centering 
    \small
    \setlength\tabcolsep{1.5pt}
    \scalebox{1}{ 
        \begin{tabular*}{\columnwidth}{@{\extracolsep{\fill}} c c c c c c @{}}
            \toprule 
             & Method & GQA & NLVR2 & Editing & Tagging\\   
            \midrule 
        
            \multirow{2}{*}{E2E} & Otter \cite{li2023otter} & 48.2 & 48.2 & - & - \\
                                 & MMICL \cite{zhao2023mmicl} & 64.4 & 62.2 & - & - \\
                                 & CFR \cite{nguyen2022coarse} & 72.1 & - & - & - \\
                                 & Qwen-vl-chat-7B \cite{bai2023qwen} & 57.5 & 58.72 & - & 32.54 \\
            \midrule
  \multirow{7}{*}{TOOL}           & Visual ChatGPT \cite{wu2023visual} & 43.2 & 51.6 & - & - \\
                                  & ViperGPT \cite{suris2023vipergpt} & 47.2 & - & - & - \\
                                  & VISPROG \cite{gupta2023visual} & 50.5 & 62.4 & 90.82 & 27.28 \\
                                  & HAMMR \cite{castrejon2024hammr} & 60.2 & 63.8 & - & - \\
                                  & ExoViP \cite{wang2024exovip} & 61.49 & 67.96 & 91.27 & 31.50 \\
                                  & CLOVA \cite{gao2024clova} & 60.2 & 63.8 & 92.16 & 31.86 \\
                                  & \textbf{Ours} & \textbf{68.4} & \textbf{68.0} & \textbf{93.44} & \textbf{31.92} \\
            \bottomrule 
        \end{tabular*}
    }
\end{table}
\subsection{Ablation Study}

We analyzed the contributions of different components within the model through ablation studies, with the results shown in Table.~\ref{ablation}. HMR significantly enhanced accuracy by managing the memory flow between the fast-access pool and the deep-retrieval pool, and continuously storing memories. TOEL further improved performance through empirical learning tailored to different task categories. By categorizing experiences and learning patterns across different categories using predefined classes and few-shot reasoning from LLMs, TOEL strengthened the model’s adaptability to diverse categories. When combined with HMR, TOEL achieved additional improvements in accuracy on both datasets. This indicates that TOEL can effectively identify and leverage category differences, thus significantly enhancing the accuracy and interpretability of the model's decision making. Additionally, the results in Table.~\ref{compare llm} further demonstrate that our approach can also significantly enhance the performance of other LLMs.

\begin{table}[htb]
\caption{\textbf{Ablation on the GQA and NLVR2 dataset.}}
\label{ablation}
\centering
\small
\setlength\tabcolsep{1.5pt}
\scalebox{1}{ 
\begin{tabular*}{\columnwidth}{@{\extracolsep{\fill}} c c c @{}}
\toprule
Dataset&Method&LLaMA2-7B\\
\midrule
\multirow{3}{*}{GQA} &Baseline & 60.2\\
                      &\textbf{+ HMR}&\textbf{64.6}\\
                      &\textbf{+ HMR + TOEL}&\textbf{68.4}\\
                      \midrule
\multirow{3}{*}{NLVR2} &Baseline & 63.8\\
                      &\textbf{+ HMR}&\textbf{65.8}\\
                      &\textbf{+ HMR + TOEL}&\textbf{68.0}\\
\bottomrule
\end{tabular*}}
\end{table}

\begin{table}[htb]
    \caption{\textbf{Comparison of results on two open-source LLMs.}} 
    \label{compare llm}
    \centering 
    \small
    \setlength\tabcolsep{1.5pt}
    \scalebox{1}{ 
       \begin{tabular*}{\columnwidth}{@{\extracolsep{\fill}} c c c c c@{}}
            \toprule 
             Method & GQA & NLVR2 & Editing & Tagging\\   
            \midrule 
             LLama2-7B + clova  &60.2 & 63.8 & 92.16 & 31.86 \\
            \textbf{LLama2-7B + ours} &\textbf{68.4} & \textbf{68.0} & \textbf{93.44} & \textbf{31.92} \\
            \midrule
            Mistral-7B + clova & 31.4 & 42.2 & 90.34 & 29.73 \\
            \textbf{Mistral-7B + Ours} & \textbf{39.6} & \textbf{43.8} & \textbf{91.82} & \textbf{29.87} \\
            \bottomrule 
        \end{tabular*}
    }
\end{table}

\subsection{Qualitative Experiment}

\begin{figure}[htb]

\begin{minipage}[b]{1.0\linewidth}
  \centering
  \centerline{\includegraphics[width=8.5cm, height=5cm, keepaspectratio]{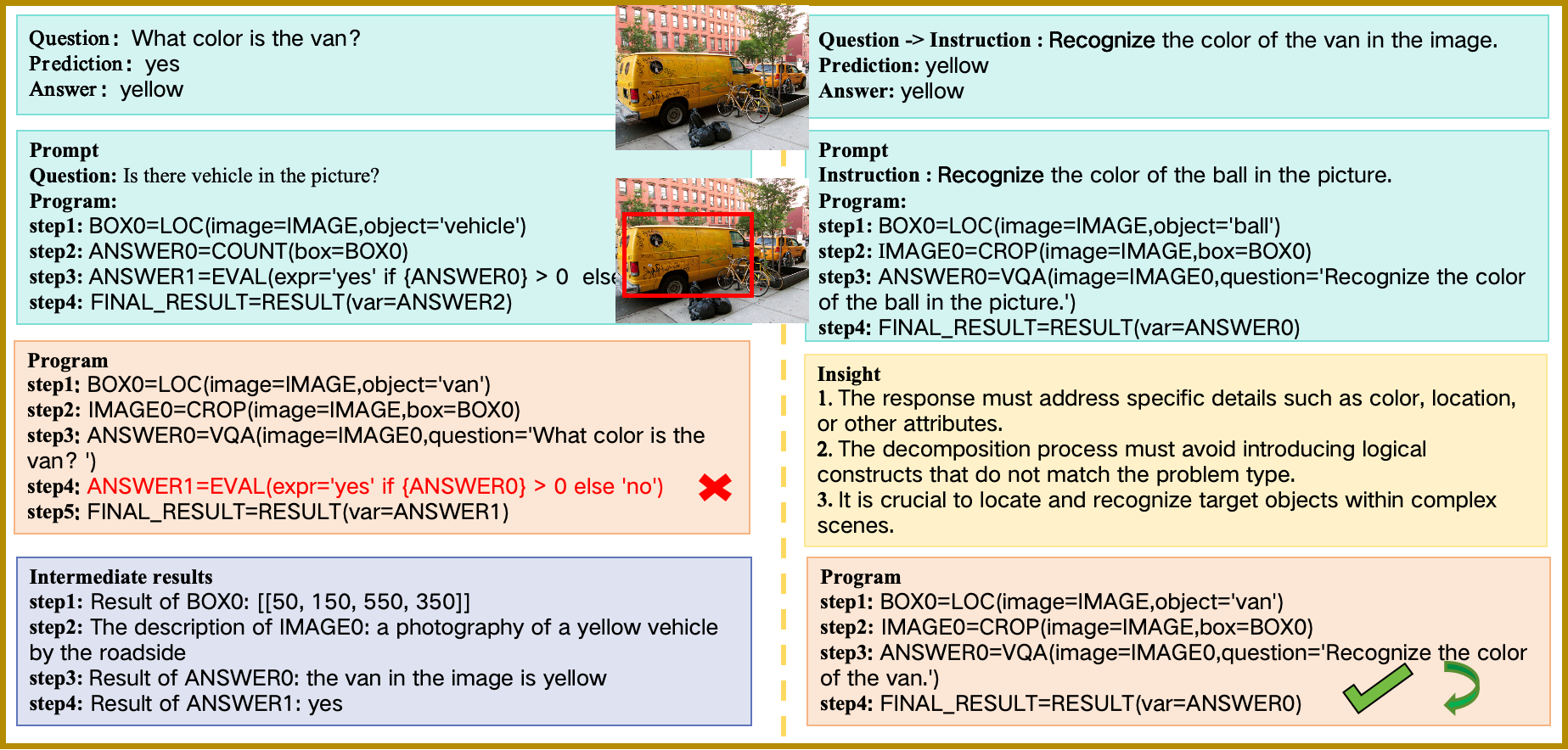}}
  \centerline{(a) Result of a compositional VQA task}\medskip
\end{minipage}

\begin{minipage}[b]{1.0\linewidth}
  \centering
  \centerline{\includegraphics[width=8.5cm, height=5cm, keepaspectratio]{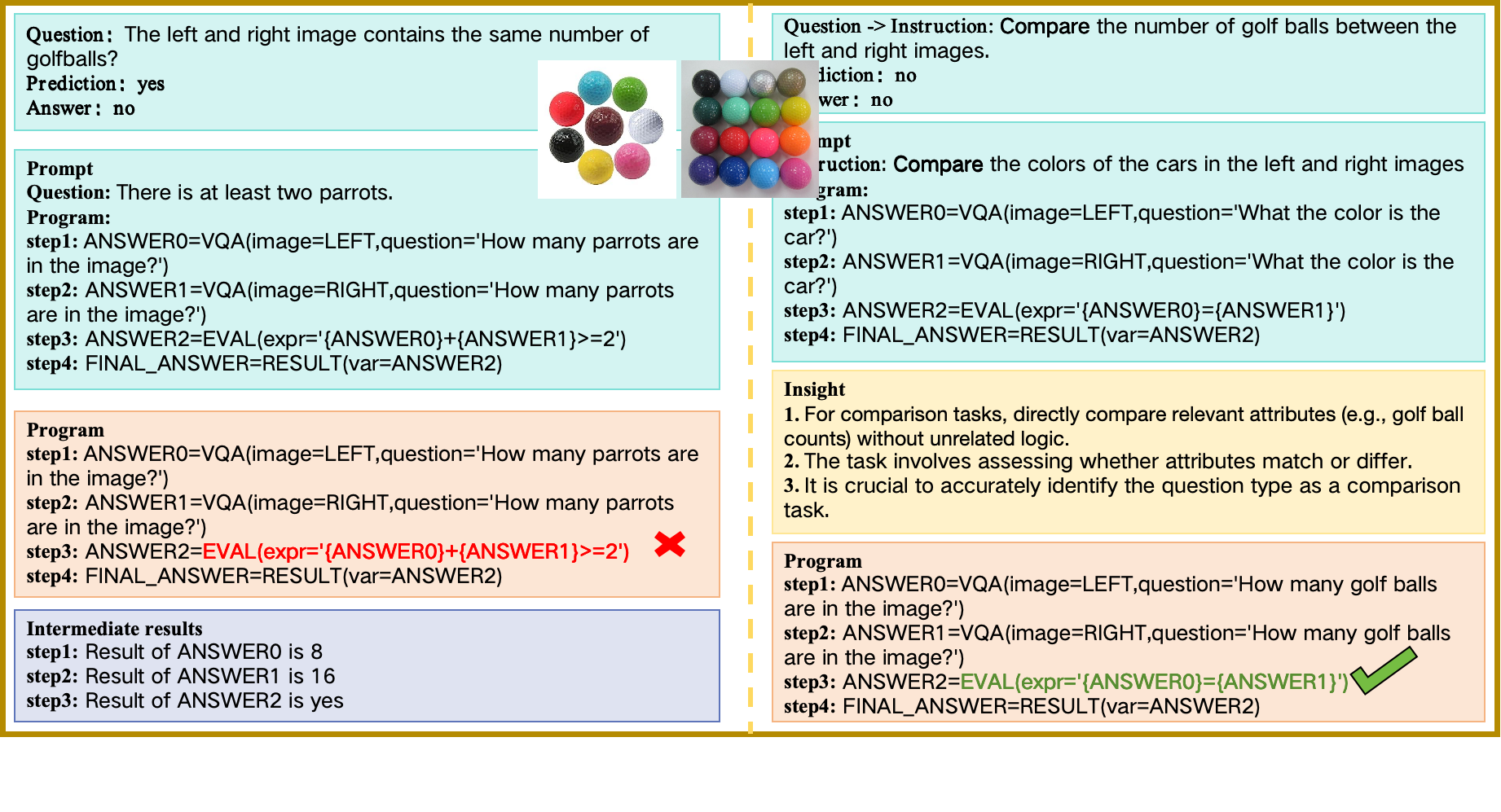}}
  \centerline{(b) Result of a multiple-image reasoning task}\medskip
\end{minipage}

\caption{Case study of EHC on two typical example tasks.}
\label{qualitive}
\end{figure}

In Fig.~\ref{qualitive}, we visualized two representative examples. The unclassified agent, unable to distinguish between categories and susceptible to interference from diverse contexts, often generates incorrect outputs. In contrast, EHC effectively enhances category understanding and selectively retrieves relevant instances, thereby ensuring category-level consistency between answers and questions.

\FloatBarrier

\section{Conclusion}

In this paper, we introduced EHC, a general agent framework that enhances decision-making capabilities through hierarchical memory management and category-based memory pattern learning. EHC consists of two core modules: Hierarchical Memory Retrieval (HMR) and Task-Oriented Experiential Learning (TOEL). Together, they enable the agent to continuously store new experiences while preserving existing knowledge during continual learning, thereby mitigating catastrophic forgetting. Simultaneously, EHC empowers the agent to learn cross-category task patterns, resulting in more accurate and interpretable decision-making. Experimental results validate the effectiveness of EHC as a general-purpose agent and highlight its strong performance and lightweight design.



\vfill
\pagebreak

\bibliographystyle{IEEEbib}
\bibliography{strings,refs}

\end{document}